\documentclass{article}

\PassOptionsToPackage{numbers, compress}{natbib}

\usepackage[preprint]{neurips_2026}

\usepackage[utf8]{inputenc} 
\usepackage[T1]{fontenc}    
\usepackage{hyperref}       
\usepackage{url}            
\usepackage{booktabs}       
\usepackage{amsfonts}       
\usepackage{nicefrac}       
\usepackage{microtype}      
\usepackage[table]{xcolor}  

\usepackage{multirow}
\usepackage{makecell}
\usepackage{arydshln}
\usepackage{bm}
\usepackage{graphicx}
\usepackage{subcaption}
\usepackage{amsmath}
\usepackage{algorithm}
\usepackage{algorithmicx}
\usepackage{algpseudocode}
\usepackage{wrapfig}

\title{Interference-Aware DeltaNet for Streaming Data}

\author{%
    Sanghyeok Chu~\footnotemark[1]~~$^{1}$ \hspace{0.6cm}
    Pyunghwan Ahn~\footnotemark[2]~~$^{2}$ \hspace{0.6cm}
    Gwangmo Song~$^{2}$ \hspace{0.6cm}
    Seung Hwan Kim~$^{2}$ \AND
    Honglak Lee~$^{2,3}$ \hspace{0.6cm}
    Bohyung Han~\footnotemark[2]~~$^{1,4}$ \\ \\
    $^1$ECE \& $^4$IPAI, Seoul National University \quad
    $^2$LG AI Research \quad
    $^3$University of Michigan \\
    \texttt{\small\{sanghyeok.chu, bhhan\}@snu.ac.kr} \\
    \texttt{\small\{p.ahn, gwangmo.song, sh.kim, honglak\}@lgresearch.ai}
}

\begin{document}

\maketitle

\renewcommand{\thefootnote}{\fnsymbol{footnote}}
\footnotetext[1]{Work done during an internship at LG AI Research.}
\footnotetext[2]{Corresponding authors.}
\renewcommand{\thefootnote}{\arabic{footnote}}


\begin{abstract}
Transformers excel at sequence modeling through attention, but their quadratic compute and growing key-value cache make them costly for long-range streaming tasks.
Linear attention variants, such as DeltaNet, address this by reformulating attention as a recurrent memory equipped with an error-corrective update rule. 
However, we show that DeltaNet's update rule induces unintended interference on past-key responses, as the correction signal leaks into every past key that is not orthogonal to the current one.
We propose Interference-Aware DeltaNet (IA-DeltaNet), which generalizes the delta rule by decoupling the key used for error evaluation from the direction along which the memory is updated.
We formulate the choice of write direction as a constrained quadratic program that minimizes the total squared interference on past-key responses while exactly preserving DeltaNet's current-key correction, and derive its closed-form solution as a preconditioned key obtained from the inverse uncentered covariance of past keys.
To ensure scalability, we further introduce a diagonal approximation of the past-key covariance, preserving the per-step efficiency and chunk-parallelizability of the original architecture.
When integrated as a memory module in a VideoLLM, IA-DeltaNet consistently outperforms baselines on streaming video understanding under equivalent bounded memory budgets, with further analyses offering insights into its memory dynamics and computational efficiency.
Collectively, these results indicate that explicitly accounting for past-key geometry is essential for designing associative memory in linear-attention architectures.

\end{abstract}

\section{Introduction}
\label{sec:introduction}

Transformers with softmax attention dominate sequence modeling, yet their quadratic compute and growing key-value cache make them costly for long or continuously arriving inputs.
Linear attention~\citep{katharopoulos2020transformers} addresses both costs by reformulating the attention as a recurrence over a fixed-size matrix-valued memory state, achieving linear-time compute and constant-memory inference. 
This recurrent formulation is particularly attractive for streaming data, where tokens arrive on the fly, the past cannot be revisited, and memory must remain bounded.

The recurrent state, however, must compress an arbitrarily long history into a finite memory, creating a tension between encoding new associations and preserving prior knowledge.
In linear attention, additive accumulation simply aggregates key-value outer products; consequently, any retrieval is inherently corrupted by interference from overlapping keys.
DeltaNet \citep{schlag2021linear} addresses this by employing an error-corrective delta-rule update.
By selectively attenuating the existing memory response at the current key before writing the new value, it achieves significantly higher fidelity in associative recall tasks.

Despite this improvement, we formally demonstrate that DeltaNet continues to suffer from inherent past-key interference.
Since at most $d_k$ directions are mutually orthogonal in a $d_k$-dimensional space, writing the correction signal along $\bm{k}_t$ inevitably perturbs overlapping past keys, leading to cumulative interference as the stream progresses.
This indicates that the fundamental limitation stems not from the error-correction mechanism itself, but from the dual role of $\bm{k}_t$ as both the read address for error evaluation and the write direction for memory updates.

To address this bottleneck, we propose Interference-Aware DeltaNet (IA-DeltaNet), a linear attention framework employing sophisticated error-corrective updates that explicitly decouple the read and write roles of the key.
While we retain $\bm{k}_t$ as the read address for evaluating prediction error, we introduce a distinct write direction $\bm{u}_t$ designed to minimize interference with past-key responses.
We formulate this as a quadratic program subject to a linear constraint that preserves the exact current-key correction of DeltaNet.
This optimization yields a closed-form solution in which an interference-aware direction is obtained by preconditioning $\bm{k}_t$ with the inverse uncentered covariance of historical keys.
Notably, our approach generalizes DeltaNet under isotropic key geometry and is theoretically guaranteed to reduce total interference compared to DeltaNet.

However, computing this exact solution requires per-step operations on a full matrix, incurring an $O(d_k^2)$ overhead even with efficient incremental updates via the Sherman--Morrison formula.
To mitigate this, we introduce a diagonal approximation that replaces the full covariance with a diagonal state, maintained as an element-wise prefix sum of squared keys.
This approximation preserves the functional form of the optimal solution by suppressing heavily-used dimensions and renormalizing to satisfy the current-key correction constraint, while adding only $O(d_k)$ complexity per step.
Furthermore, since the diagonal state can be computed as a simple prefix sum, the resulting write directions remain compatible with chunk-parallel training via associative scans.

We evaluate IA-DeltaNet in the context of streaming video understanding by integrating it as a memory module on top of a VideoLLM's vision encoder.
In this setting, the memory processes a continuous stream of visual tokens, the recurrent state remains bounded, and only the readouts from a short window of recent frames are passed to the LLM.
Our results demonstrate that IA-DeltaNet consistently outperforms established baselines across multiple benchmarks, with further analyses offering insights into its memory dynamics and computational efficiency.
Collectively, these findings suggest that explicitly accounting for past-key geometry provides a simple yet powerful principle for designing memory updates in linear-attention architectures.

Our contributions are summarized as follows:
\begin{itemize}
    \item We provide a formal analysis of memory interference in DeltaNet, demonstrating how updating along the current key direction inevitably perturbs existing associations in the recurrent state.
    \item We introduce IA-DeltaNet, which decouples the read address from the write direction within the delta-rule update, and derive an optimal write direction in closed form that minimizes past-key interference while preserving current-key fidelity. 
    \item We propose a diagonal approximation of our rule that maintains the efficiency of linear attention and enables chunk-parallel training via associative scans.
    \item We empirically validate IA-DeltaNet as a streaming memory module within a VideoLLM framework, achieving consistent performance gains over established baselines across multiple video benchmarks.
\end{itemize}


\section{Background}
\label{sec:background}

This section reviews linear attention~\citep{katharopoulos2020transformers} and DeltaNet~\citep{schlag2021linear} to establish the necessary context for analyzing memory updates and the resulting interference.

\subsection{Linear Attention as a Recurrent Memory}
\label{subsec:linear-attention}

Let $\bm{q}_t \in \mathbb{R}^{d_q}$, $\bm{k}_t \in \mathbb{R}^{d_k}$, and $\bm{v}_t\in\mathbb{R}^{d_v}$ denote the query, key, and value vectors at time step $t$, respectively. 
Linear attention~\citep{katharopoulos2020transformers} replaces the softmax operation with a linear kernel similarity. 
For simplicity, we use a dot-product kernel, $\kappa(\bm{q}_t, \bm{k}_i) \!=\! \bm{k}_i^\top \bm{q}_t$, and assume $\ell_2$-normalized keys.
By associativity, causal linear attention is expressed in the recurrent form:
\begin{equation}
    \bm{o}_t
    =
    \sum_{i\leq t}\bm{v}_i \left( \bm{k}_i^\top\bm{q}_t \right)
    =
    \left( \sum_{i\leq t}\bm{v}_i\bm{k}_i^\top \right) \bm{q}_t
    =
    \bm{S}_t\bm{q}_t,
    \qquad
    \bm{S}_t=\bm{S}_{t-1}+\bm{v}_t\bm{k}_t^\top.
\label{eq:linear-attention-recurrence}
\end{equation}
Here, $\bm{S}_t \in \mathbb{R}^{d_v \times d_k}$ is a matrix-valued memory state storing an accumulation of past key-value associations. 
This formulation decouples linear attention into a memory update and a memory readout. 
Instead of maintaining a key-value cache that grows with sequence length, the model propagates this fixed-size state, enabling constant-memory inference.

The compression inherent in this fixed-size state creates a fundamental tension between storing new information and preserving the existing memory response at the corresponding key. 
After the additive update writes the current association $\bm{v}_t \bm{k}_t^\top$, probing the memory with the current key $\bm{k}_t$ yields:
\begin{equation}
    \bm{S}_t\bm{k}_t
    = 
    (\bm{S}_{t-1}+\bm{v}_t\bm{k}_t^\top)\bm{k}_t
    =
    \bm{v}_t + \sum_{i < t} \bm{v}_i (\bm{k}_i^\top\bm{k}_t).
\label{eq:linear-current-key-response}
\end{equation}
The first term is the target value, while the second term represents the pre-existing response at $\bm{k}_t$. 
This latter term is the interference induced by previously stored values, weighted by their similarity to the current key.
Consequently, the memory response at $\bm{k}_t$ typically deviates from $\bm{v}_t$ unless the current key is orthogonal to all preceding keys.

This limitation originates from the bounded geometry of the key space. 
In $d_k$ dimensions, at most $d_k$ directions are mutually orthogonal; thus, new keys inevitably overlap with previous ones as the sequence progresses. 
Simple additive accumulation offers no mechanism to suppress this pre-existing response, leading to cumulative interference. 
This motivates memory updates that actively correct the response at the current key address rather than merely accumulating on top of the state.

\subsection{DeltaNet: Error-Corrective Update and the Resulting Past-Key Interference}
\label{subsec:deltanet}

DeltaNet~\citep{schlag2021linear} mitigates the current-key mismatch of plain additive accumulation using an error-corrective delta-rule update. 
Instead of simply accumulating the outer product $\bm{v}_t \bm{k}_t^\top$, DeltaNet first evaluates the existing memory response at the current key. 
Let $\bm{r}^{\mathrm{old}}_t := \bm{S}_{t-1}\bm{k}_t$ denote this response, and define the current-key prediction error as $\bm{e}_t := \bm{r}^{\mathrm{old}}_t - \bm{v}_t$. 
DeltaNet then applies the rank-one update:
\begin{align}
    \bm{S}_t = \bm{S}_{t-1} - \beta_t \bm{e}_t \bm{k}_t^\top = \bm{S}_{t-1} - \beta_t (\bm{r}^{\mathrm{old}}_t - \bm{v}_t) \bm{k}_t^\top,
\label{eq:deltanet-update}
\end{align}
where $\bm{o}_t = \bm{S}_t \bm{q}_t$ is the memory readout and $\beta_t$ is an input-dependent write strength. 
By expanding Eq.~\eqref{eq:deltanet-update} as $\bm{S}_t = \bm{S}_{t-1} - \beta_t (\bm{S}_{t-1}\bm{k}_t) \bm{k}_t^\top + \beta_t \bm{v}_t \bm{k}_t^\top$, it becomes clear that DeltaNet removes a fraction of the existing response at the current key and adds the target value.

To examine its effect on memory responses, let $\bm{r}^{\mathrm{new}}_j := \bm{S}_t\bm{k}_j$ and $\bm{r}^{\mathrm{old}}_j := \bm{S}_{t-1}\bm{k}_j$. 
Applying Eq.~\eqref{eq:deltanet-update} to any key $\bm{k}_j$ yields:
\begin{equation}
    \bm{r}^{\mathrm{new}}_j 
    = 
    \bm{r}^{\mathrm{old}}_j - \beta_t \bm{e}_t (\bm{k}_t^\top\bm{k}_j) 
    =
    \bm{r}^{\mathrm{old}}_j - \beta_t(\bm{k}_t^\top\bm{k}_j) \left(\bm{r}^{\mathrm{old}}_t - \bm{v}_t \right).
\label{eq:deltanet-response-change}
\end{equation}
Since $\bm{k}_t^\top\bm{k}_t=1$, setting $j=t$ yields $\bm{r}^{\mathrm{new}}_t = (1-\beta_t) \bm{r}^{\mathrm{old}}_t + \beta_t\bm{v}_t$ for the current key.
Thus, DeltaNet moves the current-key response toward the target value, allowing the new association to replace the old response rather than merely accumulate on top of it. 
This replacement behavior is particularly effective in recency-sensitive associative retrieval, where the model must return the most recently assigned value for a queried key.

However, for any past key $j < t$, Eq.~\eqref{eq:deltanet-response-change} reveals an unintended interference. 
This interference occurs because the correction signal leaks into any past key $\bm{k}_j$ that is not orthogonal to $\bm{k}_t$; the magnitude of this leakage is proportional to the key similarity $\bm{k}_t^\top \bm{k}_j$. 
Therefore, while DeltaNet successfully corrects the current-key prediction error, it fails to preserve the integrity of past-key responses. 
This observation suggests that the limitation is not the error-correction principle itself, but the lack of an interference-aware write direction that can be decoupled from the current key.

\section{Interference-Aware DeltaNet}
\label{sec:ia-deltanet}

Building upon the analysis of DeltaNet's limitations, we propose Interference-Aware DeltaNet (IA-DeltaNet). 
This approach explicitly decouples the key used for error evaluation from the direction used to update the memory.

\subsection{Main Idea}
\label{subsec:main-idea}

The core design of IA-DeltaNet is to maintain the current key $\bm{k}_t$ for measuring the prediction error, $\bm{e}_t = \bm{S}_{t-1}\bm{k}_t - \bm{v}_t$, while replacing the DeltaNet write direction $\bm{k}_t$ with an optimized interference-aware direction $\bm{u}_t$. 
The resulting update rule is defined as:
\begin{equation}
    \bm{S}_t = \bm{S}_{t-1} - \beta_t \bm{e}_t \bm{u}_t^\top,
    \qquad
    \bm{o}_t = \bm{S}_t \bm{q}_t,
\label{eq:ia-deltanet-update}
\end{equation}
where the correction signal $\bm{e}_t$, the write strength $\beta_t$, and the readout rule remain identical to those in DeltaNet. 
By changing only the direction along which the correction is written, we gain a degree of freedom to minimize memory disturbance.

Applying this update to a past key $\bm{k}_j$ ($j < t$) illustrates the impact on previous memory states:
\begin{equation}
    \bm{S}_t \bm{k}_j = \bm{S}_{t-1} \bm{k}_j - \beta_t \bm{e}_t (\bm{u}_t^\top \bm{k}_j).
\label{eq:ia-past-key-perturbation}
\end{equation}
To preserve past associations while correcting the current one, the write direction $u_t$ should satisfy two requirements:
1) it should minimize the disturbance to past-key responses, ensuring $\bm{S}_t \bm{k}_j \simeq \bm{S}_{t-1} \bm{k}_j$ for all $j < t$, and 2) it must preserve the same error-correction performance on the current key $\bm{k}_t$ as the original DeltaNet.

\subsection{Objective Function and its Closed-Form Solution}
\label{subsec:objective-and-solution}

To derive the optimal write direction $\bm{u}_t$, we first formulate the requirements introduced above using a constrained optimization framework.

\paragraph{Past-key interference objective.}

From Eq.~\eqref{eq:ia-past-key-perturbation}, the update to the current memory state induces a change in the response of every past key $\bm{k}_j$. 
We quantify the total interference as the sum of the squared $\ell_2$ norms of these interferences:
\begin{align}
    \mathcal{I}_t(\bm{u}_t) 
    = \sum_{j=1}^{t-1} \left\|-\beta_t \bm{e}_t(\bm{u}_t^\top\bm{k}_j) \right\|_2^2 
    = \beta_t^2 \|\bm{e}_t\|_2^2 \sum_{j=1}^{t-1} (\bm{u}_t^\top \bm{k}_j)^2 
    = \beta_t^2 \|\bm{e}_t\|_2^2 \bm{u}_t^\top \left( \sum_{j=1}^{t-1} \bm{k}_j\bm{k}_j^\top \right) \bm{u}_t.
\label{eq:ia-total-interference}
\end{align}
Since the write strength $\beta_t$ and the error signal $\bm{e}_t$ are independent of the choice of $\bm{u}_t$, minimizing the total interference is equivalent to minimizing the quadratic form, $\bm{u}_t^\top \bm{\Sigma}_{t-1} \bm{u}_t$. 
Here, $\bm{\Sigma}_{t-1} := \sum_{j=1}^{t-1}\bm{k}_j\bm{k}_j^\top$ represents the uncentered covariance matrix of past keys, with $\bm{\Sigma}_{0} = \bm{I}$ ensuring numerical stability.

\paragraph{Current-key correction constraint.}
While minimizing interference, we must also ensure that the update effectively corrects the response at the current key $\bm{k}_t$. 
In DeltaNet, this correction is achieved because $\bm{k}_t^\top \bm{k}_t = 1$. 
To preserve exactly the same correction behavior in IA-DeltaNet, we impose the following linear constraint:
\begin{equation}
    \bm{u}_t^\top \bm{k}_t = 1.
\label{eq:current-key-constraint}
\end{equation}
This constraint preserves the error-correction performance of DeltaNet while allowing $\bm{u}_t$ the flexibility to deviate from $\bm{k}_t$ to minimize interference with past-key responses.

\paragraph{Optimization.}

Combining the objective and the constraint, we obtain the following quadratic programming problem:
\begin{equation}
    \bm{u}_t = \underset{\hat{\bm{u}}_t}{\arg\min} ~~ \hat{\bm{u}}_t^\top \bm{\Sigma}_{t-1} \hat{\bm{u}}_t 
    \quad \text{s.t.} \quad 
    \hat{\bm{u}}_t^\top \bm{k}_t = 1.
\label{eq:ia-optimization}
\end{equation}
Using a Lagrange multiplier $\eta$, the Lagrangian is defined as
\begin{equation}
\mathcal{L}(\hat{\bm{u}}_t,\eta) = \hat{\bm{u}}_t^\top \bm{\Sigma}_{t-1} \hat{\bm{u}}_t - \eta (\hat{\bm{u}}_t^\top\bm{k}_t - 1).
\label{eq:lagrangian}
\end{equation}
The closed-form solution of this problem, representing the interference-aware write direction, is: 
\begin{equation}
    \bm{u}_t = \frac{\bm{\Sigma}_{t-1}^{-1}\bm{k}_t} {\bm{k}_t^\top \bm{\Sigma}_{t-1}^{-1} \bm{k}_t}.
\label{eq:ia-write-direction}
\end{equation}

The derived direction $\bm{u}_t$ can be interpreted as a preconditioned key. 
Specifically, $\bm{\Sigma}_{t-1}^{-1}$ acts as a whitening transform that downweights directions already heavily occupied by past keys, while the denominator renormalizes the vector to satisfy the current-key correction constraint.

IA-DeltaNet differs from DeltaNet when the past-key geometry is anisotropic; in the isotropic case ($\bm{\Sigma}_{t-1} \!\propto\! \bm{I}$), the optimal write direction reduces to $\bm{u}_t \!=\! \bm{k}_t$, recovering DeltaNet.
Thus, IA-DeltaNet generalizes DeltaNet by adapting the write direction to how past keys are distributed across the representation space.

\subsection{Geometric Interpretation of Interference Reduction}
\label{subsec:ia-interpretation}

Although $|\bm{u}_t^\top \bm{k}_j| < |\bm{k}_t^\top \bm{k}_j|$ does not hold for every individual past key, IA-DeltaNet is mathematically guaranteed to reduce the total interference. 
Let $\mathcal{I}_t(\bm{w}) = \bm{w}^\top \bm{\Sigma}_{t-1} \bm{w}$ be the total squared interference. 
For DeltaNet ($\bm{u}_t = \bm{k}_t$), the interference is $\bm{k}_t^\top \bm{\Sigma}_{t-1} \bm{k}_t$. For IA-DeltaNet, substituting the optimal solution from Eq.~\eqref{eq:ia-write-direction} yields:
\begin{equation}
    \mathcal{I}_t(\bm{u}_t) = \frac{1}{\bm{k}_t^\top \bm{\Sigma}_{t-1}^{-1} \bm{k}_t}.
\label{eq:ia-minimized-interference}
\end{equation}
By applying the matrix-form Cauchy-Schwarz inequality, specifically $(\bm{k}_t^\top \bm{\Sigma}_{t-1} \bm{k}_t)(\bm{k}_t^\top \bm{\Sigma}_{t-1}^{-1} \bm{k}_t) \geq (\bm{k}_t^\top \bm{k}_t)^2 = 1$, we have the following inequality:
\begin{equation}
    \mathcal{I}_t(\bm{u}_t) \leq \mathcal{I}_t(\bm{k}_t),
\end{equation}
where the equality holds when $\bm{k}_t$ is an eigenvector of $\bm{\Sigma}_{t-1}$, meaning that the current key already aligns with the principal directions of past interference. 
In all other cases, IA-DeltaNet strictly reduces the total memory disturbance by "whitening" the write direction through $\bm{\Sigma}_{t-1}^{-1}$, effectively steering the update away from high-density regions of the past-key geometry.

\subsection{Efficient Diagonal Approximation}
\label{subsec:diagonal-ia-deltanet}

While the exact solution in Eq.~\eqref{eq:ia-write-direction} provides the optimal interference-aware direction, it poses computational challenges for high-dimensional states. 
A direct implementation requires either a full matrix inversion or a linear system solve at each step. 
Although the inverse covariance can be updated via the Sherman--Morrison formula by exploiting the recursive structure, $\bm{\Sigma}_t = \bm{\Sigma}_{t-1} + \bm{k}_t \bm{k}_t^\top$, this approach still incurs an $O(d_k^2)$ cost per step. 
To maintain the linear-time efficiency characteristic of modern recurrent models, we adopt a diagonal approximation of the covariance matrix.

Specifically, we approximate the uncentered covariance matrix $\bm{\Sigma}_{t-1}$ with a diagonal matrix:
\begin{equation}
    \bm{\Sigma}_{t-1} \approx \bm{D}_{t-1} = \operatorname{diag}(\bm{d}_{t-1}),
\end{equation}
where the diagonal state $\bm{d}_t \!\in\! \mathbb{R}^{d_k}$ is computed as a prefix sum of elementwise squared keys:
\begin{equation}
    \bm{d}_t = \bm{d}_{t-1} + \bm{k}_t \odot \bm{k}_t, \quad \bm{d}_0 = \bm{1}.
\label{eq:diagonal-state-update}
\end{equation}
Substituting $\bm{D}_{t-1}$ into the solution in Eq.~\eqref{eq:ia-write-direction} yields the interference-aware write direction:
\begin{equation}
    \bm{u}_t = \frac{\bm{D}_{t-1}^{-1}\bm{k}_t}{\bm{k}_t^\top \bm{D}_{t-1}^{-1}\bm{k}_t}.
\label{eq:diagonal-write-direction}
\end{equation}

In this formulation, the diagonal state $\bm{d}_{t-1}$ serves as an accumulator of feature-wise energy from past keys. 
Consequently, this approximation selectively downweights dimensions that have been heavily occupied by previous keys, then renormalizes the direction to ensure that the current-key correction constraint is strictly satisfied.


\vspace{-2mm}
\begin{algorithm}[t]
\caption{Diagonal IA-DeltaNet in recurrent form.}
\label{alg:diagonal-ia-deltanet}
\begin{algorithmic}[1]
\Require keys $\{\bm{k}_t\}_{t=1}^T$, queries $\{\bm{q}_t\}_{t=1}^T$, values $\{\bm{v}_t\}_{t=1}^T$, write strengths $\{\beta_t\}_{t=1}^T$
\State Initialize $\bm{S}_0\gets\bm{0}\in\mathbb{R}^{d_v\times d_k}$
\State Initialize $\bm{d}_0\gets\bm{1}\in\mathbb{R}^{d_k}$
\For{$t=1,\ldots,T$}
    \State $\bm{D}_{t-1} \gets \operatorname{diag}(\bm{d}_{t-1})$
    \State $\bm{u}_t \gets \bm{D}_{t-1}^{-1} \bm{k}_t / (\bm{k}_t^\top \bm{D}_{t-1}^{-1} \bm{k}_t)$ \Comment{IA write direction, Eq.~\eqref{eq:diagonal-write-direction}}
    \State $\bm{e}_t \gets \bm{S}_{t-1}\bm{k}_t-\bm{v}_t$
    \State $\bm{S}_t \gets \bm{S}_{t-1}-\beta_t\bm{e}_t\bm{u}_t^\top$ \Comment{IA-DeltaNet update, Eq.~\eqref{eq:ia-deltanet-update}}
    \State $\bm{o}_t \gets \bm{S}_t\bm{q}_t$
    \State $\bm{d}_t \gets \bm{d}_{t-1}+\bm{k}_t\odot\bm{k}_t$ \Comment{Update diagonal state, Eq.~\eqref{eq:diagonal-state-update}}
\EndFor
\State \Return $\{\bm{o}_t\}_{t=1}^T$
\end{algorithmic}
\end{algorithm}

The resulting recurrent procedure is summarized in Algorithm~\ref{alg:diagonal-ia-deltanet}. 
The computational overhead relative to the standard DeltaNet is minimal, consisting only of the element-wise construction of $\bm{u}_t$ and the update of $\bm{d}_t$. 
Crucially, the dominant operation remains the $O(d_v d_k)$ rank-one memory update; since the additional overhead for computing $\bm{u}_t$ is only $O(d_k)$, IA-DeltaNet maintains the same asymptotic complexity as the original DeltaNet.
Moreover, since $\bm{d}_t$ can be computed by a simple prefix sum, the diagonal write directions can be computed efficiently in a chunk-parallel fashion using associative scans, preserving the efficiency of linear attention.

\section{Experiments}
\subsection{Experimental Setup}
\label{sec:experimental_setup}
\paragraph{Architecture and baselines.}
We build on Qwen3.5-4B~\citep{qwen35blog} and insert memory blocks between the vision aligner and the LLM. 
These memory blocks provide cross-frame information flow before the visual tokens are passed to the LLM, propagating temporal information through their memory state. 
We stack two memory blocks, each consisting of a temporal-mixing layer followed by an FFN. 
We compare three categories of temporal-mixing mechanisms: (1) memory layers that compress past frames into a fixed-sized state $\bm{S}$, including DeltaNet~\citep{schlag2021linear}, GatedDeltaNet~\citep{yanggated}, and our IA-DeltaNet; (2) causal attention~\citep{vaswani2017attention}, which retains a full key-value cache rather than compressing into a fixed-size state; and (3) an identity baseline, where the temporal mixer is replaced with an identity mapping.
Note that IA-DeltaNet refers to its diagonally-approximated variant throughout our experiments.

\vspace{-2mm}
\paragraph{Streaming protocol.}
We evaluate all models in a streaming setup designed to isolate the role of memory.
The memory blocks process the video stream in temporal order, continuously updating their recurrent state across frames. The LLM then receives only the readout from the last $W$ frames; earlier frames propagate to the LLM only through the memory state.
This contrasts with the standard offline setting, where all frames are directly visible to the LLM.

\vspace{-2mm}

\paragraph{Benchmarks.}
We evaluate models on a suite of video QA benchmarks covering three evaluation goals.
First, VNBench~\citep{zhaoneedle} is a video needle-in-the-haystack benchmark that directly measures how well a model preserves and retrieves past information from memory. 
Second, TempCompass~\citep{liu2024tempcompass}, NExT-QA~\citep{xiao2021next}, MVBench~\citep{li2024mvbench}, and TemporalBench~\citep{cai2024temporalbench} assess temporal and motion reasoning across frames.
Finally, Video-MME~\citep{fu2025video} provides an evaluation across diverse real-world video understanding tasks.
We select these benchmarks since their accuracy changes meaningfully with the visible-window size $W$, indicating that solving them requires memory to recover broader video context beyond the visible local window.
Appendix~\ref{appx:benchmark_choice} details the visible-window analysis used for benchmark selection.

\vspace{-2mm}
\paragraph{Training and evaluation.}
We train models on LLaVA-Video-178K~\citep{lin2024video} for $1$ epoch using MS-SWIFT~\citep{zhao2025swift}. 
The vision encoder and aligner are frozen, while the LLM is adapted with LoRA using rank $8$ and scaling factor $\alpha=16$. 
Training uses $16$ H200 GPUs with an effective batch size of $64$. 
We optimize with AdamW ($\beta_1 \!=\! 0.9$, $\beta_2 \!=\! 0.95$) using learning rates of $2 \!\times\! 10^{-4}$ for the memory blocks and $2 \!\times\! 10^{-5}$ for the LoRA parameters, with a cosine decay schedule and a $3\%$ warmup ratio.
For both training and evaluation, videos are sampled at $1$ FPS with a frame-token budget of $16$ to $128$.
We cap each video at $180$ frames during training. 
We pass the memory readouts from the last $W \!=\! 8$ frames to the LLM during training; at evaluation, we report results for multiple values of $W$.


\begin{figure}[t]
    \centering
    \vspace{-2mm}
    \includegraphics[width=\linewidth]{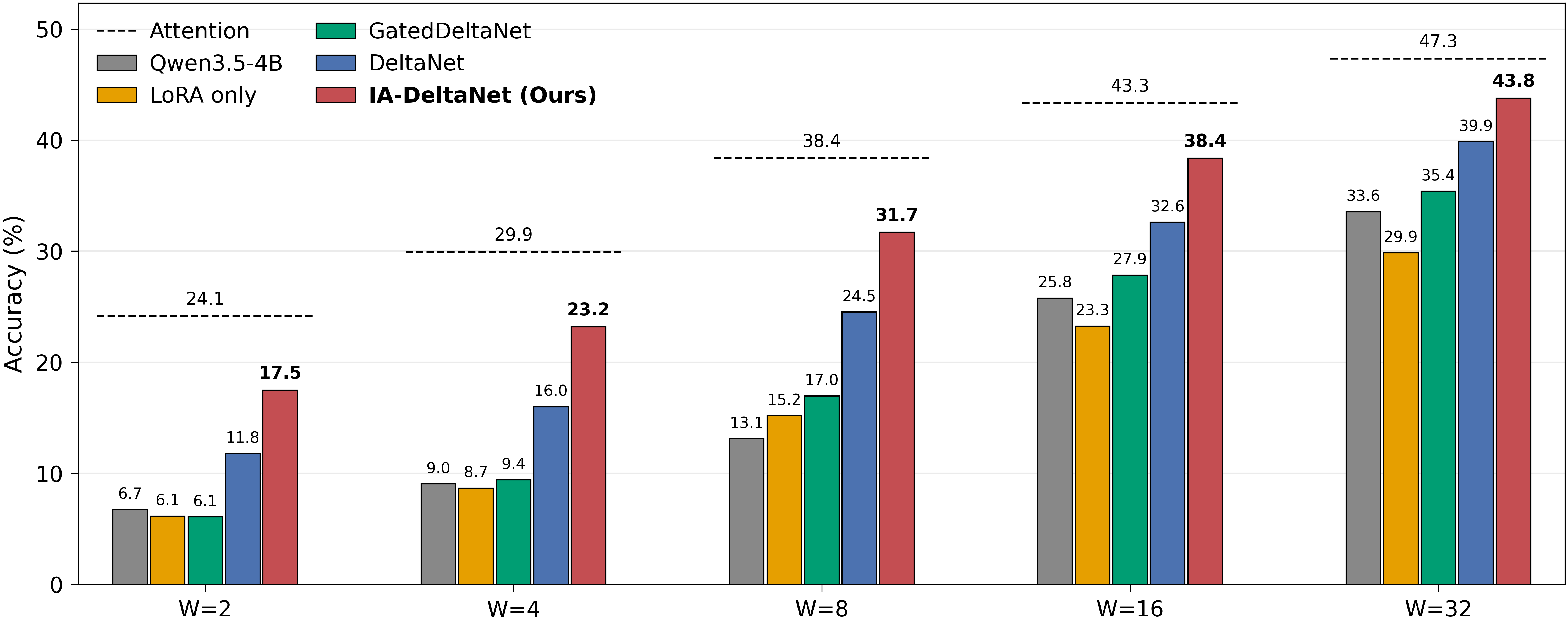}
    \caption{
    VNBench~\citep{zhaoneedle} accuracy across LLM-visible window sizes $W$. 
    Dashed lines indicate the attention, which requires retaining all past KVs.
    IA-DeltaNet consistently outperforms fixed-size recurrent memory baselines and progressively narrows the gap to attention as $W$ grows.
    }
    \label{fig:vnbench_sweep}
    \vspace{-1mm}
\end{figure}

\begin{table}[t]
    \centering
    \renewcommand{\arraystretch}{1.05}
    \setlength{\tabcolsep}{7.5pt}
    \vspace{-2mm}
    \caption{
    Results on TempCompass~\citep{liu2024tempcompass}, NExT-QA~\citep{xiao2021next}, MVBench~\citep{li2024mvbench}, and TemporalBench~\citep{cai2024temporalbench}.
    Identity denotes the LoRA-only baseline with temporal mixer replaced with an identity mapping.
    Bold indicates the best performance among models with fixed-size state.
    $\dagger$ denotes full-KV reference.}
    \label{table:lmms-eval}
    \scalebox{0.85}{
    \begin{tabular}{lcccccc}
    \toprule
    Model & Memory & Cost & TempCompass & NExT-QA & MVBench & TemporalBench \\
    \midrule
    \rowcolor{gray!15} 
    \multicolumn{7}{c}{Last $W \!=\! 2$ frames} \\
    Qwen3.5-4B & \multirow{2}{*}{No mem.} & \multirow{2}{*}{$\mathcal{O}(1)$} & 54.4 & 67.4 & 51.6 & 55.0 \\
    + Identity & & & 59.5 & 66.4 & 51.9 & 61.4 \\
    \cdashline{1-7}[0.4pt/1.2pt]
    + DeltaNet~\citep{schlag2021linear} & \multirow{3}{*}{State $\bm{S}$} & \multirow{3}{*}{$\mathcal{O}(1)$} & 59.4 & 60.0 & \textbf{55.2} & 62.0 \\
    + GatedDeltaNet~\citep{yanggated} & & & 58.6 & 67.5 & 52.4 & 61.1 \\
    + \textbf{IA-DeltaNet (Ours)} & & & \textbf{65.4} & \textbf{71.1} & \textbf{55.2} & \textbf{63.5} \\
    \cdashline{1-7}[0.4pt/1.2pt]
    + Attention$^{\dagger}$~\citep{vaswani2017attention} & Full KV & $\mathcal{O}(N)$ & 63.2 & 71.7 & 56.7 & 64.4 \\
    \midrule
    \rowcolor{gray!15} 
    \multicolumn{7}{c}{Last $W \!=\! 4$ frames} \\
    Qwen3.5-4B & \multirow{2}{*}{No mem.} & \multirow{2}{*}{$\mathcal{O}(1)$} & 60.0 & 70.2 & 57.0 & 56.9 \\
    + Identity & & & 62.9 & 69.1 & 54.5 & 63.1 \\
    \cdashline{1-7}[0.4pt/1.2pt]
    + DeltaNet~\citep{schlag2021linear} & \multirow{3}{*}{State $\bm{S}$} & \multirow{3}{*}{$\mathcal{O}(1)$} & 62.6 & 66.2 & 57.7 & 64.5 \\
    + GatedDeltaNet~\citep{yanggated} & & & 62.7 & 70.0 & 56.3 & 64.2 \\
    + \textbf{IA-DeltaNet (Ours)} & & & \textbf{67.7} & \textbf{75.0} & \textbf{58.4} & \textbf{64.6} \\
    \cdashline{1-7}[0.4pt/1.2pt]
    + Attention$^{\dagger}$~\citep{vaswani2017attention} & Full KV & $\mathcal{O}(N)$ & 66.8 & 75.4 & 57.6 & 65.8 \\
    \midrule
    \rowcolor{gray!15} 
    \multicolumn{7}{c}{Last $W \!=\! 8$ frames} \\
    Qwen3.5-4B & \multirow{2}{*}{No mem.} & \multirow{2}{*}{$\mathcal{O}(1)$} & 66.3 & 72.7 & 58.7 & 58.0 \\
    + Identity & & & 64.9 & 71.9 & 55.9 & 64.2 \\
    \cdashline{1-7}[0.4pt/1.2pt]
    + DeltaNet~\citep{schlag2021linear} & \multirow{3}{*}{State $\bm{S}$} & \multirow{3}{*}{$\mathcal{O}(1)$} & 66.4 & 73.5 & 58.3 & 64.6 \\
    + GatedDeltaNet~\citep{yanggated} & & & 67.5 & 73.6 & 58.1 & \textbf{65.4} \\
    + \textbf{IA-DeltaNet (Ours)} & & & \textbf{70.1} & \textbf{76.9} & \textbf{59.2} & \textbf{65.4} \\
    \cdashline{1-7}[0.4pt/1.2pt]
    + Attention$^{\dagger}$~\citep{vaswani2017attention} & Full KV & $\mathcal{O}(N)$ & 68.8 & 76.7 & 58.1 & 65.7 \\
    \bottomrule
    \end{tabular}}
    \vspace{-1mm}
\end{table}

\subsection{Quantitative Results}
\subsubsection{Video Needle in the Haystack}
Figure~\ref{fig:vnbench_sweep} reports VNBench accuracy as the number of LLM-visible frames $W$ varies.
Across all $W$, no-memory baselines are weak by construction because the needle-frame information is likely outside the visible window, while attention provides a strong reference by retaining the full KV cache.
Among recurrent memory models, IA-DeltaNet consistently performs best across all values of $W$.
The advantage is most pronounced in the small-$W$ regime, where the model must rely most heavily on the recurrent state to recover needle-frame information.
GatedDeltaNet remains close to the no-memory baselines in this regime, likely because global forgetting can remove needle information during subsequent updates.
As $W$ increases, IA-DeltaNet narrows the gap to attention, from $6.6\%$ points at $W \!=\! 2$ to $3.5\%$ points at $W \!=\! 32$.
Overall, the results support that interference-aware writes improve finite-state memory retention.

\subsubsection{Temporal and Motion-centric Video QA}
Table~\ref{table:lmms-eval} evaluates temporal and motion-centric video QA benchmarks, which complement VNBench by testing whether memory benefits carry over to reasoning over information distributed across frames.
Similar to VNBench, IA-DeltaNet consistently outperforms other bounded-memory models across different window sizes, with the largest advantage at $W \!=\! 2$.
As $W$ increases, the gap narrows, but IA-DeltaNet remains the best bounded-memory model across benchmarks.
Despite compressing past frames into a fixed-size state, IA-DeltaNet remains competitive with attention on several benchmarks.
These results suggest that interference-aware writes improve memory-efficient temporal reasoning beyond the controlled retrieval setting of VNBench.


\begin{wraptable}{r}{0.28\linewidth}
\vspace{-6mm}
\centering
\renewcommand{\arraystretch}{1.05}
\setlength{\tabcolsep}{2pt}
\caption{Quantitative results on Video-MME~\citep{fu2025video}.}
\label{table:videomme}
\resizebox{\linewidth}{!}{
\begin{tabular}{lc}
\toprule
Model & Video-MME \\
\midrule
Qwen3.5-4B & 49.7 \\
+ Identity & 50.4 \\
\cdashline{1-2}[0.4pt/1.2pt]
+ DeltaNet~\citep{schlag2021linear} & 53.5 \\
+ GatedDeltaNet~\citep{yanggated} & 51.4 \\
+ \textbf{IA-DeltaNet} & \textbf{54.9} \\
\cdashline{1-2}[0.4pt/1.2pt]
+ Attention$^{\dagger}$~\citep{vaswani2017attention} & 58.4 \\
\bottomrule
\end{tabular}}
\vspace{-4mm}
\end{wraptable}

\subsubsection{General Video QA on Video-MME}
We further evaluate general video understanding on Video-MME, which contains relatively long videos with diverse lengths and covers a broader range of video QA skills.
We set $W$ to $8$, $64$, and $128$ for the short, medium, and long subsets, respectively.
As shown in Table~\ref{table:videomme}, IA-DeltaNet again performs best among memory models with a fixed-size state, showing that interference-aware writes generalize beyond specialized memory or temporal reasoning benchmarks.
At the same time, the gap to attention is larger on Video-MME, likely because its longer videos are harder to compress into a fixed-size state.

\subsection{Analysis}
\label{appx:analysis}
%

\begin{figure}[t]
    \centering
    \vspace{-2mm}
    \begin{subfigure}[t]{0.32\linewidth}
        \centering
        \includegraphics[width=\linewidth]{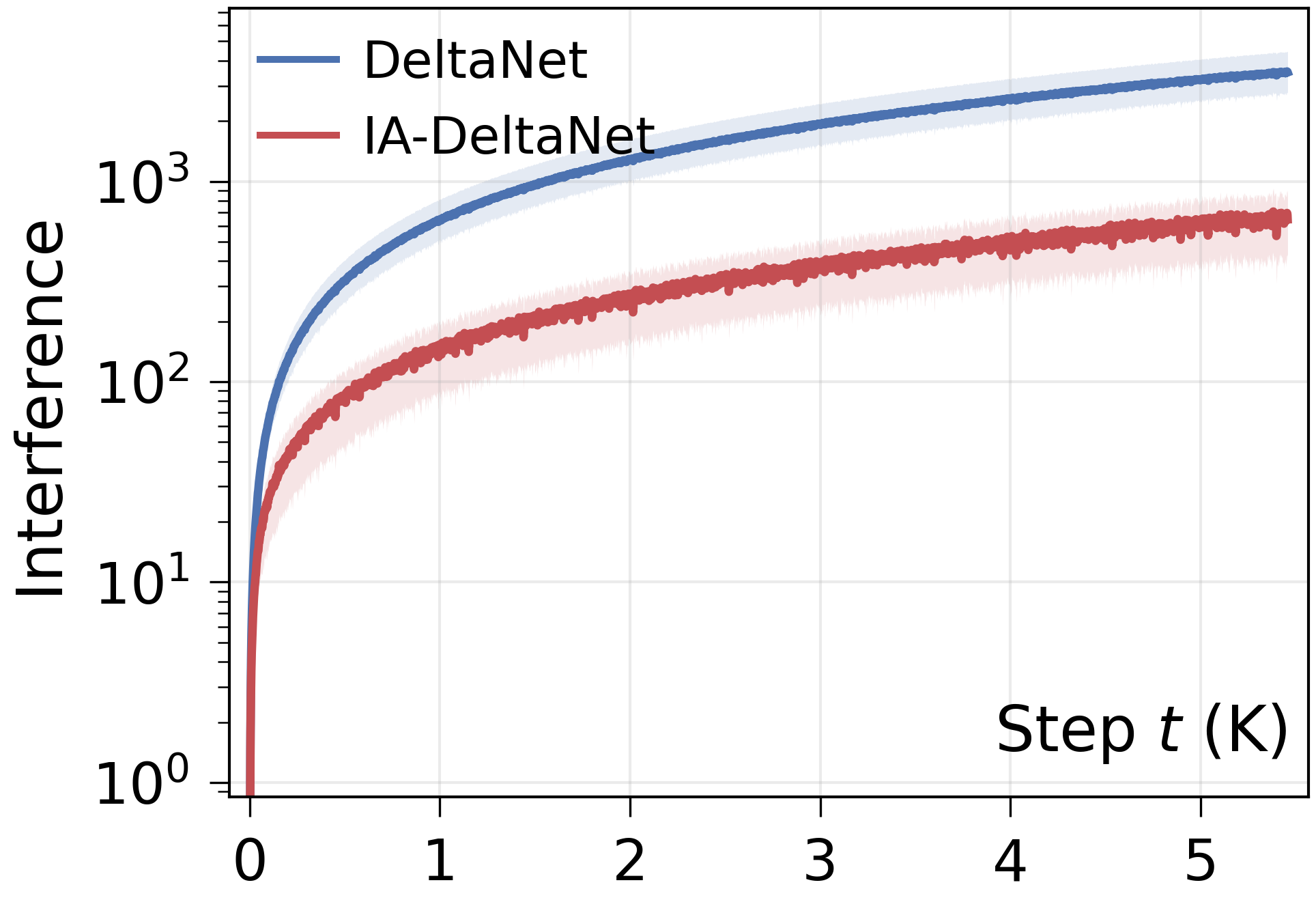}
        \caption{Interference.}
        \label{fig:interference_real}
    \end{subfigure}
    \hfill
    \begin{subfigure}[t]{0.32\linewidth}
        \centering
        \includegraphics[width=\linewidth]{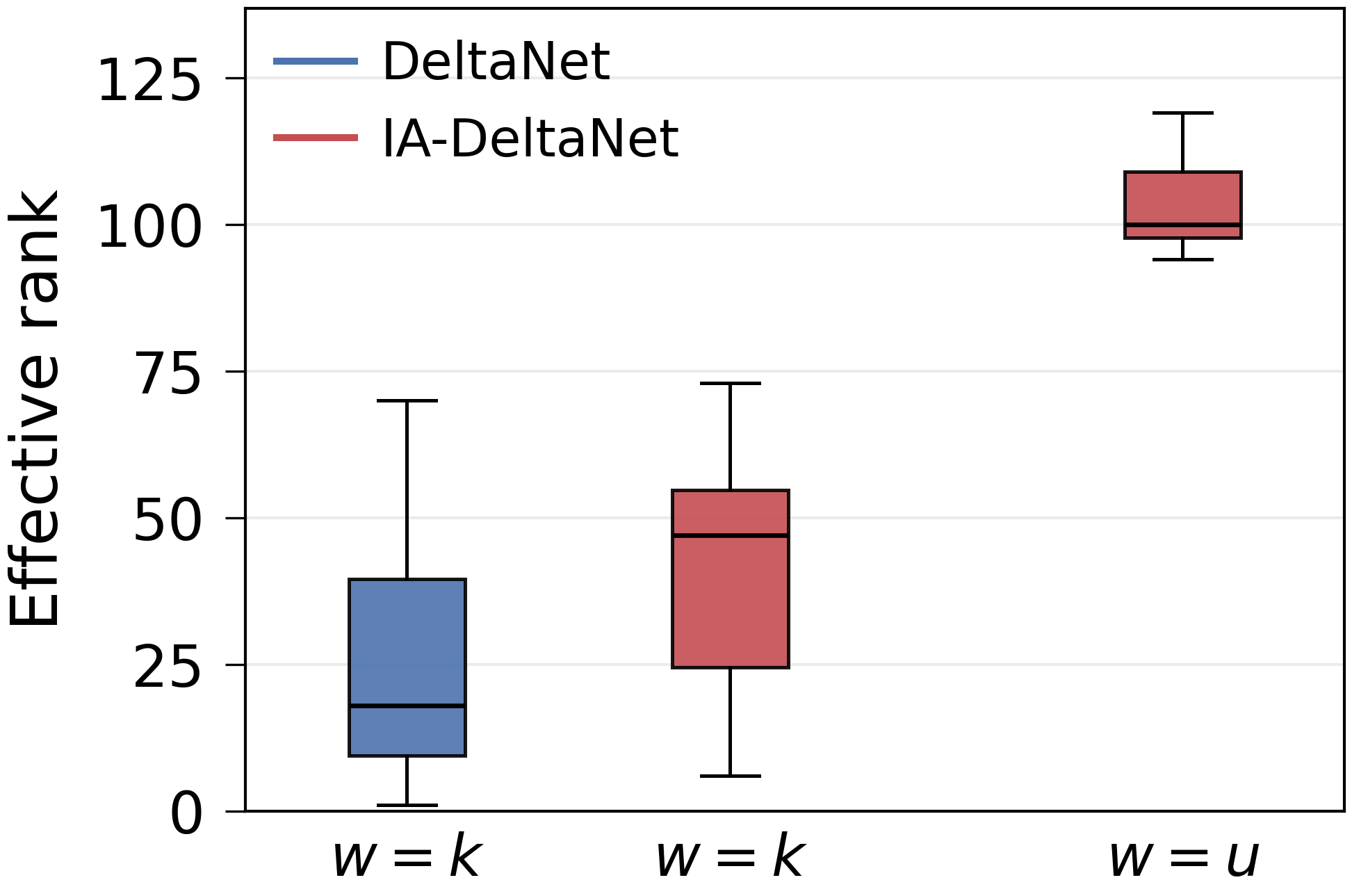}
        \caption{Effective Rank}
        \label{fig:analysis_rank}
    \end{subfigure}
    \hfill
    \begin{subfigure}[t]{0.32\linewidth}
        \centering
        \includegraphics[width=\linewidth]{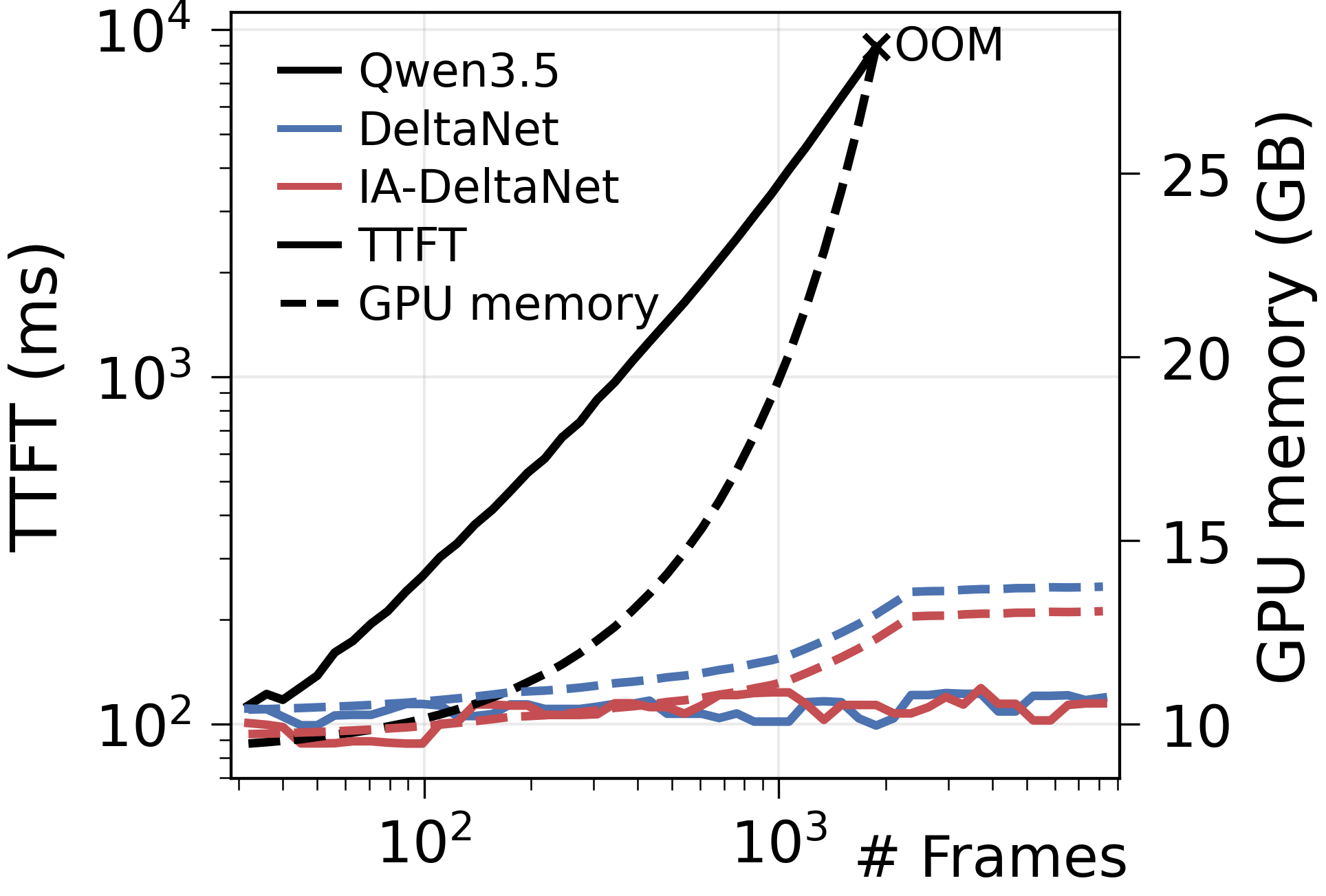}
        \caption{Efficiency Comparison.}
        \label{fig:analysis_efficiency}
    \end{subfigure}
    \caption{
	Analysis of IA-DeltaNet. 
        (a) IA-DeltaNet reduces write interference on real video keys compared with DeltaNet. 
        (b) IA-DeltaNet increases the effective rank of learned keys and write directions.
        (c) IA-DeltaNet preserves the efficiency of fixed-size recurrent memory.
        }
    \label{figure:analysis}
    \vspace{-1mm}
\end{figure}

\subsubsection{Interference of Memory Write}
\label{subsubsec:analysis_interference}
We measure the interference of a write direction $\bm{w}$ as $\mathcal{I}_t(\bm{w}) \!=\! \bm{w}^\top \bm{\Sigma}_{t-1} \bm{w}$, which quantifies its overlap with the accumulated past-key covariance.
We compare DeltaNet ($\bm{w} \!=\! \bm{k}_t$), exact IA-DeltaNet ($\bm{w} \!=\! \bm{u}_t$), and its diagonal approximation ($\bm{w} \!=\! \bm{u}_t^{\mathrm{diag}}$) on both synthetic keys and real video keys.

\paragraph{Synthetic keys.}
As established in Section~\ref{subsec:objective-and-solution}, IA-DeltaNet is equivalent to DeltaNet under isotropic past-key distributions and is theoretically guaranteed to reduce interference for anisotropic cases.
This raises a question regarding the diagonal approximation: it becomes less faithful when the past-key covariance $\bm{\Sigma}_{t-1}$ contains non-zero off-diagonal entries, but such correlations simultaneously imply an anisotropic structure that can increase unintended interference in the standard delta rule. 
Since analytically comparing these competing effects---diagonal approximation error of IA-DeltaNet versus DeltaNet's increased interference under anisotropic key geometry---is non-trivial, we empirically evaluate them by drawing synthetic keys $\bm{k}_t \!=\! \tilde{\bm{k}}_t / \|\tilde{\bm{k}}_t\|_2$ from a controlled distribution, where $\tilde{\bm{k}}_t \!\sim\! \mathcal{N}(\bm{0}, \bm{Q}(\theta)\bm{\Lambda}\bm{Q}(\theta)^\top)$.
By sweeping across varying degrees of anisotropy ($\bm{\Lambda}$) and eigenspace rotations ($\theta$), we observe that the interference of DeltaNet serves as an empirical upper bound for that of the diagonal IA-DeltaNet.
This suggests that a simple diagonal preconditioner can still reduce interference sufficiently to maintain a consistent advantage over DeltaNet across diverse geometric configurations.
Detailed experimental setups and full results are provided in Appendix~\ref{sec:appendix_synthetic_keys}.

\paragraph{Real video keys.}
In addition to the synthetic case, we observed that the same trend holds for real video inputs.
On the Video-MME medium subset, we compute $\mathcal{I}_t(\bm{w})$ at each timestep using keys extracted from each trained model.
As shown in Figure~\ref{fig:interference_real}, IA-DeltaNet yields lower interference than DeltaNet across timesteps, mirroring the synthetic results.
Together, these analyses show that the interference-reducing behavior of diagonal IA-DeltaNet observed in synthetic settings also carries over to real video keys from trained models.

\subsubsection{Effective Rank of Write Directions}
To better understand how each model uses the write-direction space, we measure the effective rank of write-related directions per video on the Video-MME medium subset.
As shown in Figure~\ref{fig:analysis_rank}, DeltaNet keys $\bm{k}$ have a median effective rank of $17$ out of $d_k \!=\! 160$, and IA-DeltaNet keys $\bm{k}$ have a higher median rank of $49$, indicating a broader learned key subspace.
The IA-DeltaNet write directions $\bm{u}$, computed from $\bm{k}$, further increase the median rank to $96$, as expected from their interference-aware construction.

\subsubsection{Efficiency Comparison}
Figure~\ref{fig:analysis_efficiency} reports time-to-first-token (TTFT) and GPU memory usage. 
For Qwen3.5, the measurement includes LLM prefill and first-token decoding with the accumulated visual tokens as input. 
For DeltaNet and IA-DeltaNet, it includes memory-block processing followed by LLM prefill and first-token decoding with the memory readouts as input.
Unlike Qwen3.5, whose latency and memory grow with the number of accumulated visual tokens and eventually lead to out-of-memory on longer inputs, DeltaNet and IA-DeltaNet compress the video stream into a fixed-size recurrent state.
As a result, their memory usage and TTFT remain much more stable as the number of frames increases.
IA-DeltaNet adds only a small overhead over DeltaNet on both measures.


\section{Related Work}
\label{sec:related}

\paragraph{Linear attention as recurrent memory.}
Linear attention reformulates softmax attention as a recurrent update over a matrix-valued state~\citep{katharopoulos2020transformers}.
DeltaNet improves this formulation by replacing additive accumulation with an error-corrective delta-rule update~\citep{schlag2021linear}, and subsequent recurrent memory models have explored chunk-parallel training, gating mechanisms, more expressive hidden states, and test-time memorization~\citep{yang2024parallelizing,yanggated,sun2025learning,behrouztitans}.
These methods typically write new information along the current key direction.
We revisit this design choice and derive an interference-aware write direction that preserves the current key correction while minimally perturbing previously stored associations.

\paragraph{Memory efficient long-context understanding.}
A broad line of work addresses long-context modeling by retaining, compressing, or retrieving a bounded subset of past information.
StreamingLLM maintains attention sinks and recent tokens for stable streaming inference~\citep{xiaoefficient}, while H$_2$O, SnapKV, and PyramidKV reduce the KV cache by selecting important past tokens under a fixed memory budget~\citep{zhang2023h2o, li2024snapkv, cai2025pyramidkv}.
For video understanding, long-form and streaming VideoLLMs further reduce memory pressure through frame sampling, token compression, memory banks, sparse long-term memories, or video-specific KV-cache mechanisms~\citep{lin2024video,chen2025longvila,he2024malmm,song2024moviechat,flashvstream,qian2024streaming,di2025rekv}.
These methods primarily decide which past information to retain, compress, or retrieve from an explicit cache.

\paragraph{Sub-quadratic attention.}
A large body of work seeks to reduce the quadratic cost of self-attention through kernel approximations~\citep{katharopoulos2020transformers, choromanski2021rethinking}, low-rank or sparse attention patterns~\citep{wang2020linformer,beltagy2020longformer,zaheer2020bigbird}, and recurrent or state-space sequence models~\citep{guefficiently, mamba, mamba2, yanggated}.
While these approaches differ in formulation, they share the goal of processing long sequences with sub-quadratic computation or bounded-size states.


\section{Conclusion}

We presented IA-DeltaNet, an interference-aware recurrent memory update that decouples the key used for error evaluation from the direction along which memory is written.
This write direction minimizes past-key interference while preserving the current-key correction.
We further introduced a diagonal covariance approximation that retains linear-time efficiency and chunk-parallel training with minimal overhead.
Integrated into Qwen3.5, IA-DeltaNet outperforms fixed-size recurrent memory baselines across multiple benchmarks, under bounded memory budgets.
Our analyses show that IA-DeltaNet reduces write interference on real video keys, uses a higher-rank write-direction space, and preserves the efficiency of fixed-size recurrent memory.
These results suggest that explicitly accounting for past-key geometry is a simple yet effective design principle for memory updates in linear-attention architectures.

\clearpage
{
    \small
    \bibliographystyle{plainnat}
    \bibliography{neurips_2026}

@String(CVPR= {IEEE Conf. Comput. Vis. Pattern Recog.})

@String(ICCV= {Int. Conf. Comput. Vis.})

@String(NIPS= {Adv. Neural Inform. Process. Syst.})

@String(ICASSP=	{ICASSP})

@String(ICLR = {Int. Conf. Learn. Represent.})

@String(AAAI = {AAAI})

@String(CVPR  = {CVPR})

@String(ICCV  = {ICCV})

@String(NIPS  = {NeurIPS})

@String(ICLR  = {ICLR})

@inproceedings{zhaoneedle,
  title={Needle In A Video Haystack: A Scalable Synthetic Evaluator for Video MLLMs},
  author={Zhao, Zijia and Lu, Haoyu and Huo, Yuqi and Du, Yifan and Yue, Tongtian and Guo, Longteng and Wang, Bingning and Liu, Jing and others},
  booktitle={ICLR},
  year={2025}
}

@inproceedings{fu2025video,
  title={Video-MME: The first-ever comprehensive evaluation benchmark of multi-modal llms in video analysis},
  author={Fu, Chaoyou and Dai, Yuhan and Luo, Yongdong and Li, Lei and Ren, Shuhuai and Zhang, Renrui and Wang, Zihan and Zhou, Chenyu and Shen, Yunhang and Zhang, Mengdan and others},
  booktitle={CVPR},
  year={2025}
}

@inproceedings{liu2024tempcompass,
  title={TempCompass: Do video llms really understand videos?},
  author={Liu, Yuanxin and Li, Shicheng and Liu, Yi and Wang, Yuxiang and Ren, Shuhuai and Li, Lei and Chen, Sishuo and Sun, Xu and Hou, Lu},
  booktitle={ACL Findings},
  year={2024}
}

@inproceedings{xiao2021next,
  title={NExT-QA: Next phase of question-answering to explaining temporal actions},
  author={Xiao, Junbin and Shang, Xindi and Yao, Angela and Chua, Tat-Seng},
  booktitle={CVPR},
  year={2021}
}

@inproceedings{li2024mvbench,
  title={MVBench: A comprehensive multi-modal video understanding benchmark},
  author={Li, Kunchang and Wang, Yali and He, Yinan and Li, Yizhuo and Wang, Yi and Liu, Yi and Wang, Zun and Xu, Jilan and Chen, Guo and Luo, Ping and others},
  booktitle={CVPR},
  year={2024}
}

@inproceedings{cai2024temporalbench,
  title={TemporalBench: Benchmarking fine-grained temporal understanding for multimodal video models},
  author={Cai, Mu and Tan, Reuben and Zhang, Jianrui and Zou, Bocheng and Zhang, Kai and Yao, Feng and Zhu, Fangrui and Gu, Jing and Zhong, Yiwu and Shang, Yuzhang and others},
  journal={NeurIPS, workshop on Video-Language Models},
  year={2024}
}

@inproceedings{schlag2021linear,
  title={Linear transformers are secretly fast weight programmers},
  author={Schlag, Imanol and Irie, Kazuki and Schmidhuber, J{\"u}rgen},
  booktitle={ICML},
  year={2021},
}

@inproceedings{yanggated,
  title={Gated Delta Networks: Improving Mamba2 with Delta Rule},
  author={Yang, Songlin and Kautz, Jan and Hatamizadeh, Ali},
  booktitle={ICLR},
  year={2025},
}

@inproceedings{behrouztitans,
  title={Titans: Learning to Memorize at Test Time},
  author={Behrouz, Ali and Zhong, Peilin and Mirrokni, Vahab},
  booktitle={NeurIPS},
  year={2025}
}

@inproceedings{katharopoulos2020transformers,
  title={Transformers are RNNS: Fast autoregressive transformers with linear attention},
  author={Katharopoulos, Angelos and Vyas, Apoorv and Pappas, Nikolaos and Fleuret, Fran{\c{c}}ois},
  booktitle={ICML},
  year={2020}
}

@misc{qwen35blog,
    title={Qwen3.5: Towards Native Multimodal Agents},
    url={https://qwen.ai/blog?id=qwen3.5},
    author={QwenTeam},
    year={2026}
}

@article{vaswani2017attention,
  title={Attention is all you need},
  author={Vaswani, Ashish and Shazeer, Noam and Parmar, Niki and Uszkoreit, Jakob and Jones, Llion and Gomez, Aidan N and Kaiser, {\L}ukasz and Polosukhin, Illia},
  journal={NIPS},
  year={2017}
}

@inproceedings{zhao2025swift,
  title={Swift: a scalable lightweight infrastructure for fine-tuning},
  author={Zhao, Yuze and Huang, Jintao and Hu, Jinghan and Wang, Xingjun and Mao, Yunlin and Zhang, Daoze and Jiang, Zeyinzi and Wu, Zhikai and Ai, Baole and Wang, Ang and others},
  booktitle={AAAI},
  year={2025}
}

@inproceedings{
    choromanski2021rethinking,
    title={Rethinking Attention with Performers},
    author={Krzysztof Marcin Choromanski and Valerii Likhosherstov and David Dohan and Xingyou Song and Andreea Gane and Tamas Sarlos and Peter Hawkins and Jared Quincy Davis and Afroz Mohiuddin and Lukasz Kaiser and David Benjamin Belanger and Lucy J Colwell and Adrian Weller},
    booktitle={ICLR},
    year={2021},
}

@article{wang2020linformer,
  title={Linformer: Self-attention with linear complexity},
  author={Wang, Sinong and Li, Belinda Z and Khabsa, Madian and Fang, Han and Ma, Hao},
  booktitle={arXiv},
  year={2020}
}

@article{beltagy2020longformer,
  title={Longformer: The long-document transformer},
  author={Beltagy, Iz and Peters, Matthew E and Cohan, Arman},
  journal={arXiv},
  year={2020}
}

@inproceedings{zaheer2020bigbird,
  title={Big bird: Transformers for longer sequences},
  author={Zaheer, Manzil and Guruganesh, Guru and Dubey, Kumar Avinava and Ainslie, Joshua and Alberti, Chris and Ontanon, Santiago and Pham, Philip and Ravula, Anirudh and Wang, Qifan and Yang, Li and others},
  booktitle={NeurIPS},
  year={2020}
}

@inproceedings{guefficiently,
  title={Efficiently Modeling Long Sequences with Structured State Spaces},
  author={Gu, Albert and Goel, Karan and Re, Christopher},
  booktitle={ICLR},
  year={2022}
}

@article{mamba,
  title={Mamba: Linear-Time Sequence Modeling with Selective State Spaces},
  author={Gu, Albert and Dao, Tri},
  journal={arXiv},
  year={2023}
}

@inproceedings{mamba2,
  title={Transformers are {SSM}s: Generalized Models and Efficient Algorithms Through Structured State Space Duality},
  author={Dao, Tri and Gu, Albert},
  booktitle={ICML},
  year={2024}
}

@article{yang2024parallelizing,
  title={Parallelizing linear transformers with the delta rule over sequence length},
  author={Yang, Songlin and Wang, Bailin and Zhang, Yu and Shen, Yikang and Kim, Yoon},
  journal={NeurIPS},
  year={2024}
}

@inproceedings{xiaoefficient,
  title={Efficient Streaming Language Models with Attention Sinks},
  author={Xiao, Guangxuan and Tian, Yuandong and Chen, Beidi and Han, Song and Lewis, Mike},
  booktitle={ICLR},
  year={2024}
}

@article{zhang2023h2o,
  title={H2o: Heavy-hitter oracle for efficient generative inference of large language models},
  author={Zhang, Zhenyu and Sheng, Ying and Zhou, Tianyi and Chen, Tianlong and Zheng, Lianmin and Cai, Ruisi and Song, Zhao and Tian, Yuandong and R{\'e}, Christopher and Barrett, Clark and others},
  journal={NeurIPS},
  year={2023}
}

@article{li2024snapkv,
  title={Snapkv: Llm knows what you are looking for before generation},
  author={Li, Yuhong and Huang, Yingbing and Yang, Bowen and Venkitesh, Bharat and Locatelli, Acyr and Ye, Hanchen and Cai, Tianle and Lewis, Patrick and Chen, Deming},
  journal={NeurIPS},
  year={2024}
}

@inproceedings{
    cai2025pyramidkv,
    title={Pyramid{KV}: Dynamic {KV} Cache Compression based on Pyramidal Information Funneling},
    author={Zefan Cai and Yichi Zhang and Bofei Gao and Yuliang Liu and Yucheng Li and Tianyu Liu and Keming Lu and Wayne Xiong and Yue Dong and Junjie Hu and Wen Xiao},
    booktitle={CoLM},
    year={2025},
}

@inproceedings{lin2024video,
  title={{Video-LLaVA}: Learning united visual representation by alignment before projection},
  author={Lin, Bin and Ye, Yang and Zhu, Bin and Cui, Jiaxi and Ning, Munan and Jin, Peng and Yuan, Li},
  booktitle={EMNLP},
  year={2024}
}

@inproceedings{
    chen2025longvila,
    title={Long{VILA}: Scaling Long-Context Visual Language Models for Long Videos},
    author={Yukang Chen and Fuzhao Xue and Dacheng Li and Qinghao Hu and Ligeng Zhu and Xiuyu Li and Yunhao Fang and Haotian Tang and Shang Yang and Zhijian Liu and Yihui He and Hongxu Yin and Pavlo Molchanov and Jan Kautz and Linxi Fan and Yuke Zhu and Yao Lu and Song Han},
    booktitle={ICLR},
    year={2025},
}

@inproceedings{he2024malmm,
  title={{MA-LMM}: Memory-Augmented Large Multimodal Model for Long-Term Video Understanding},
  author={He, Bo and Li, Hengduo and Jang, Young Kyun and Jia, Menglin and Cao, Xuefei and Shah, Ashish and Shrivastava, Abhinav and Lim, Ser-Nam},
  booktitle={CVPR},
  year={2024}
}

@inproceedings{song2024moviechat,
  title={Moviechat: From dense token to sparse memory for long video understanding},
  author={Song, Enxin and Chai, Wenhao and Wang, Guanhong and Zhang, Yucheng and Zhou, Haoyang and Wu, Feiyang and Chi, Haozhe and Guo, Xun and Ye, Tian and Zhang, Yanting and others},
  booktitle={CVPR},
  year={2024}
}

@inproceedings{flashvstream,
  title={Flash-VStream: Efficient Real-Time Understanding for Long Video Streams},
  author={Zhang, Haoji and Wang, Yiqin and Tang, Yansong and Liu, Yong and Feng, Jiashi and Dai, Jifeng and Jin, Xiaojie},
  booktitle={ICCV},
  year={2025}
}

@article{qian2024streaming,
  title={Streaming long video understanding with large language models},
  author={Qian, Rui and Dong, Xiaoyi and Zhang, Pan and Zang, Yuhang and Ding, Shuangrui and Lin, Dahua and Wang, Jiaqi},
  journal={NeurIPS},
  year={2024}
}

@inproceedings{di2025rekv,
  title={Streaming Video Question-Answering with In-context Video KV-Cache Retrieval},
  author={Di, Shangzhe and Yu, Zhelun and Zhang, Guanghao and Li, Haoyuan and Cheng, Hao and Li, Bolin and He, Wanggui and Shu, Fangxun and Jiang, Hao and others},
  booktitle={ICLR},
  year={2025}
}

@inproceedings{
    sun2025learning,
    title={Learning to (Learn at Test Time): {RNN}s with Expressive Hidden States},
    author={Yu Sun and Xinhao Li and Karan Dalal and Jiarui Xu and Arjun Vikram and Genghan Zhang and Yann Dubois and Xinlei Chen and Xiaolong Wang and Sanmi Koyejo and Tatsunori Hashimoto and Carlos Guestrin},
    booktitle={ICML},
    year={2025},
}

@inproceedings{lin2026streamingbench,
  title={{StreamingBench}: Assessing the gap for mllms to achieve streaming video understanding},
  author={Lin, Junming and Fang, Zheng and Chen, Chi and Cheng, Haoxuan and Wan, Zihao and Luo, Fuwen and Wang, Ziyue and Li, Peng and Liu, Yang and Sun, Maosong},
  booktitle={ICASSP},
  year={2026},
}

@inproceedings{niu2025ovo,
  title={{OVO-Bench}: How far is your video-llms from real-world online video understanding?},
  author={Niu, Junbo and Li, Yifei and Miao, Ziyang and Ge, Chunjiang and Zhou, Yuanhang and He, Qihao and Dong, Xiaoyi and Duan, Haodong and Ding, Shuangrui and Qian, Rui and others},
  booktitle={CVPR},
  year={2025}
}
}

\clearpage
\appendix

\setcounter{section}{0}
\renewcommand{\thesection}{\Alph{section}}
\renewcommand{\thefigure}{\Alph{section}.\arabic{figure}}
\counterwithin{figure}{section}
\renewcommand{\thetable}{\Alph{section}.\arabic{table}}
\counterwithin{table}{section}
\renewcommand{\theequation}{\Alph{section}.\arabic{equation}}
\counterwithin{equation}{section}

\appendix

\section{Evaluation Details and Additional Results}
\label{appx:evaluation_details}

\subsection{Benchmark Choice and Justification}
\label{appx:benchmark_choice}

\begin{figure}[H]
    \centering
    \includegraphics[width=0.9\linewidth]{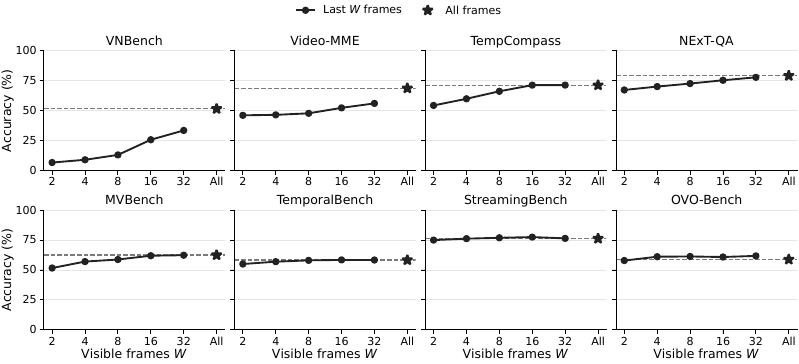}
    \vspace{-1mm}
    \caption{
    Visible-window sensitivity of Qwen3.5.
    Benchmarks are sorted by the accuracy gain from $W=2$ to $W=180$.
    Benchmarks with larger gains are used in the main evaluation because they provide a clearer test of memory beyond a short visible window.
    }
    \vspace{-1mm}
    \label{fig:appendix_benchmark_analysis}
\end{figure}

\paragraph{Visible-window sensitivity analysis.}
To assess whether a benchmark is suitable for evaluating memory under our streaming protocol, we analyze how accuracy changes as the number of LLM-visible frames $W$ varies.
If accuracy improves substantially with larger $W$, the benchmark is likely to require information beyond the last few visible frames, making it a clearer test of memory.
We sample each video at $1$ FPS and provide only the last $W$ frames to Qwen3.5-4B~\citep{qwen35blog}; for videos longer than three minutes, we uniformly sample $180$ frames before selecting the last $W$ frames.
Figure~\ref{fig:appendix_benchmark_analysis} reports accuracy curves across different values of $W$, ranging from $2$ to $180$.

\paragraph{Video QA benchmarks.}
As shown in Figure~\ref{fig:appendix_benchmark_analysis}, the evaluated video QA benchmarks generally improve as the visible window $W$ increases.
VNBench~\citep{zhaoneedle} shows the largest gain as $W$ increases, as expected from its needle-in-the-haystack design.
Video-MME~\citep{fu2025video}, TempCompass~\citep{liu2024tempcompass}, NExT-QA~\citep{xiao2021next}, TemporalBench~\citep{cai2024temporalbench}, and MVBench~\citep{li2024mvbench} also generally improve with larger visible windows, indicating that they benefit from information beyond a short local window.
Although these benchmarks were originally designed for offline video QA, this visible-window sensitivity supports their use for evaluating memory under our streaming protocol.

\paragraph{Streaming QA benchmarks.}
Streaming-oriented benchmarks such as StreamingBench~\citep{lin2026streamingbench} and OVO-Bench~\citep{niu2025ovo} follow a query-time evaluation protocol, where each question is answered using the video stream observed up to its query time rather than the entire video.
Accordingly, for our visible-window analysis, we sample frames at $1$ FPS up to each query time and provide only the last $W$ frames to Qwen3.5.
At $W \!=\! 2$, the drop relative to $W \!=\! 180$ is only $1.3\%$ points on StreamingBench and $0.9\%$ points on OVO-Bench.
At $W \!=\! 8$, using only the last $8$ frames even outperforms the $W \!=\! 180$ setting on both StreamingBench ($76.9\%$ vs. $76.2\%$) and OVO-Bench ($61.3\%$ vs. $58.8\%$).
This suggests that, under their query-time protocol, a short last-$W$ window already provides sufficient evidence for these benchmarks, making them less suitable for evaluating whether a recurrent state preserves information over the full video stream.
For this reason, we focus our main evaluation on benchmarks whose performance depends more clearly on information beyond a short local window.

\subsection{Model Configuration}
\label{sec:appendix_model_config}

Table~\ref{table:model_config} summarizes the additional trainable parameters introduced by each memory block type.
The identity baseline contains only the shared FFN and normalization parameters.
All memory block variants add only a small number of trainable parameters relative to the $4$B-parameter Qwen3.5 backbone.
DeltaNet and IA-DeltaNet have the same parameter count because IA-DeltaNet changes only the write direction without adding trainable parameters.
GatedDeltaNet has more parameters because we follow its default configuration, which doubles the value dimension $d_v$.


\begin{table}[h]
    \centering
    \renewcommand{\arraystretch}{1.05}
    \setlength{\tabcolsep}{11pt}
    \caption{
    Additional trainable parameters for each memory block type.
    Identity denotes the LoRA-only baseline where the temporal mixer is replaced with an identity mapping.
    }
    \vspace{1mm}
    \label{table:model_config}
    \scalebox{0.95}{
    \begin{tabular}{lc}
    \toprule
    Memory block & Parameters \\
    \midrule
    Identity & $53.1$M \\
    DeltaNet & $79.4$M \\
    GatedDeltaNet & $105.6$M \\
    \textbf{IA-DeltaNet (Ours)} & $79.4$M \\
    \bottomrule
    \end{tabular}
    }
\end{table}

\section{Details on Write Interference Analysis}
\label{appx:analysis_details}
This section provides additional details for the write interference analysis in Section~\ref{subsubsec:analysis_interference}.
We first describe the synthetic key distribution used to control covariance anisotropy and eigenspace rotation, and then present the full interference sweep over these factors.

\subsection{Synthetic Key Distribution}
\label{sec:appendix_synthetic_keys}
We construct synthetic keys to control the geometry of the past-key covariance.
The construction separates the eigenvalue spectrum, which determines anisotropy, from the eigenvector basis, whose rotation determines the amount of off-diagonal covariance.
Specifically, we first draw raw keys $\tilde{\bm{k}}_t \!\sim\! \mathcal{N}(\bm{0}, \bm{C}(\theta))$, and then obtain the normalized keys as $\bm{k}_t \!=\! \tilde{\bm{k}}_t / \|\tilde{\bm{k}}_t\|_2$.
We construct the covariance as $\bm{C}(\theta) \!=\! \bm{Q}(\theta)\bm{\Lambda}\bm{Q}(\theta)^\top$.
Here, $\bm{\Lambda} \!=\! \operatorname{diag}(\lambda_1,\dots,\lambda_{d_k})$ sets the anisotropic eigenvalue spectrum, where $\lambda_i \!=\! \exp(-\alpha(i-1))$ for $i=1,\dots,d_k$.
The decay rate $\alpha$ controls the eigenvalue decay and therefore the effective rank of the key covariance.

To construct the eigenvector matrix $\bm{Q}(\theta)$, we first define a random matrix $\bm{A}$ that determines the rotation direction.
For each trial, we sample a matrix $\bm{R}$ with entries $R_{ij}\sim\mathcal{N}(0,1)$ and define
\begin{equation}
	\bm{A}=\bm{R}-\bm{R}^\top,
	\qquad
	\bm{A}\leftarrow \frac{\bm{A}}{\|\bm{A}\|_F}\sqrt{d_k}.
\end{equation}
We then define $\bm{Q}(\theta)=\exp(\theta\bm{A})$, which is orthogonal by construction since $\bm{A}^\top=-\bm{A}$.
Thus, the columns of $\bm{Q}(\theta)$ form an orthonormal eigenbasis for $\bm{C}(\theta)$.
Let $\bm{q}_i$ denote the $i$-th column of $\bm{Q}(\theta)$.
Since $\bm{Q}(\theta)$ is orthogonal, $\bm{Q}(\theta)^\top \bm{q}_i=\bm{e}_i$.
Thus,
\begin{equation}
	\bm{C}(\theta)\bm{q}_i = \bm{Q}(\theta)\bm{\Lambda}\bm{Q}(\theta)^\top \bm{q}_i = \lambda_i\bm{q}_i.
\end{equation}
The parameter $\theta$ controls how far this eigenbasis is rotated from the coordinate axes.
At $\theta \!=\! 0$, $\bm{Q}(0) \!=\! \bm{I}$ and the covariance is diagonal.
As $\theta$ increases, the same eigenvalue spectrum is preserved, but the eigenspace is rotated away from the coordinate axes, introducing off-diagonal covariance and making the diagonal approximation less faithful.
The same $\bm{A}$ is reused for all values of $\theta$ within a trial, so the sweep follows a fixed rotation path.

\subsection{Synthetic Interference Sweep}
\label{sec:appendix_synthetic_sweep}
We evaluate the interference measure $\mathcal{I}_t(\bm{w})$ for DeltaNet, exact IA-DeltaNet, and diagonal IA-DeltaNet under the synthetic key setup defined above.
We use $d_k \!=\! 160$, sequence length $T \!=\! 5000$, and average over $15$ independent trials per setting.
In this sweep, we evaluate $\theta \!\in\! \{0, \pi/32, \pi/16, \pi/8, \pi/4\}$ in radians and set $\alpha$ to match target effective ranks $r_{\mathrm{eff}} \!\in\! \{5,\,10,\,20,\,80,\,160\}$.

\begin{figure}[t]
    \centering
    \includegraphics[width=\linewidth]{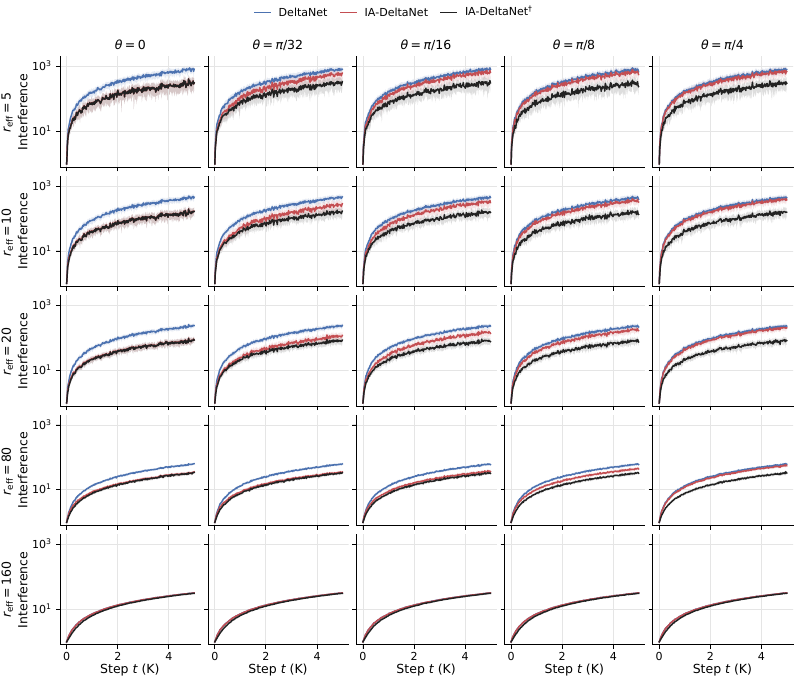}
    \caption{
    Synthetic interference sweep over covariance rotation $\theta$ and anisotropy levels ($r_{\text{eff}}$).
    The diagonal approximation closely follows the exact IA-DeltaNet update when the covariance is diagonal or mildly rotated, while the gap increases as off-diagonal covariance becomes stronger.
    }
    \label{fig:synthetic_interference_sweep}
\end{figure}

Figure~\ref{fig:synthetic_interference_sweep} provides the full sweep over covariance rotation and anisotropy level.
The diagonal approximation most closely follows exact IA-DeltaNet when the covariance is diagonal or only mildly rotated.
As $\theta$ increases, off-diagonal covariance grows, and the gap between diagonal and exact IA-DeltaNet becomes larger.
Across anisotropy levels, interference-aware write directions are most beneficial when the past-key covariance is highly anisotropic, corresponding to low effective-rank settings where keys concentrate in a smaller subspace and key-tied DeltaNet writes induce larger interference.
As the covariance becomes more diffuse and closer to isotropic, DeltaNet, exact IA-DeltaNet, and diagonal IA-DeltaNet behave more similarly.

\section{Algorithms and Pseudocode}
\label{appx:pseudo}

\subsection{Algorithm for IA-DeltaNet in Recurrent Form}
\label{appx:psuedo-recurrent}

Algorithm~\ref{alg:ia-deltanet-recurrent} presents the recurrent form of the original IA-DeltaNet without the diagonal approximation.
For reference, Algorithm~\ref{alg:deltanet-recurrent} also shows the corresponding DeltaNet recurrence.
Compared to DeltaNet, IA-DeltaNet computes the interference-aware write direction $\bm{u}_t$ from Eq.~\eqref{eq:ia-write-direction} and updates the overlap state $\bm{\Sigma}_t$.
The memory update then replaces the write-side key $\bm{k}_t$ with $\bm{u}_t$.


%
\begin{algorithm}[h]
\caption{IA-DeltaNet in recurrent form.}
\label{alg:ia-deltanet-recurrent}
\begin{algorithmic}[1]
\Require keys $\{\bm{k}_t\}_{t=1}^T$, queries $\{\bm{q}_t\}_{t=1}^T$, values $\{\bm{v}_t\}_{t=1}^T$, gates $\{\beta_t\}_{t=1}^T$
\State $\bm{S}_0 \gets \bm{0}\in\mathbb{R}^{d_v\times d_k}$
\State $\bm{\Sigma}_0 \gets \bm{I}\in\mathbb{R}^{d_k\times d_k}$
\For{$t=1,\ldots,T$}
    \State $\bm{z}_t \gets \bm{\Sigma}_{t-1}^{-1}\bm{k}_t$
    \State $\bm{u}_t \gets \bm{z}_t / (\bm{k}_t^{\!\top}\bm{z}_t)$ \Comment{IA write direction Eq.~\eqref{eq:ia-write-direction}}
    \State $\bm{e}_t \gets \bm{S}_{t-1}\bm{k}_t - \bm{v}_t$
    \State $\bm{S}_t \gets \bm{S}_{t-1} - \beta_t\,\bm{e}_t\,\bm{u}_t^{\!\top}$
    \State $\bm{\Sigma}_t \gets \bm{\Sigma}_{t-1} + \bm{k}_t\bm{k}_t^{\!\top}$
    \State $\bm{o}_t \gets \bm{S}_t\bm{q}_t$
\EndFor
\State \Return $\{\bm{o}_t\}_{t=1}^T$
\end{algorithmic}
\end{algorithm}
\begin{algorithm}[h]
\caption{DeltaNet in recurrent form.}
\label{alg:deltanet-recurrent}
\begin{algorithmic}[1]
\Require keys $\{\bm{k}_t\}_{t=1}^T$, queries $\{\bm{q}_t\}_{t=1}^T$, values $\{\bm{v}_t\}_{t=1}^T$, gates $\{\beta_t\}_{t=1}^T$
\State $\bm{S}_0 \gets \bm{0}\in\mathbb{R}^{d_v\times d_k}$
\For{$t=1,\ldots,T$}
    \State $\bm{e}_t \gets \bm{S}_{t-1}\bm{k}_t - \bm{v}_t$
    \State $\bm{S}_t \gets \bm{S}_{t-1} - \beta_t\,\bm{e}_t\,\bm{k}_t^{\!\top}$
    \State $\bm{o}_t \gets \bm{S}_t\bm{q}_t$
\EndFor
\State \Return $\{\bm{o}_t\}_{t=1}^T$
\end{algorithmic}
\end{algorithm}

\subsection{Sherman--Morrison Rank-1 Inverse Update}
\label{appx:pseudo-sm}

The exact recurrence in Algorithm~\ref{alg:ia-deltanet-recurrent} requires inverting $\bm{\Sigma}_{t-1}$ at every step.
Because $\bm{\Sigma}_t = \bm{\Sigma}_{t-1} + \bm{k}_t\bm{k}_t^{\!\top}$ is a rank-$1$ update, the inverse can be maintained directly via Sherman--Morrison.
Let
\begin{equation}
\label{eq:sm-pieces}
    \bm{p}_t \;=\; \bm{\Sigma}_{t-1}^{-1}\bm{k}_t,
    \qquad
    s_t \;=\; \bm{k}_t^{\!\top}\bm{p}_t.
\end{equation}
Then,
\begin{equation}
\label{eq:sm-update}
    \bm{\Sigma}_t^{-1} \;=\; \bm{\Sigma}_{t-1}^{-1} \;-\; \frac{\bm{p}_t \bm{p}_t^{\!\top}}{1 + s_t},
\end{equation}
This avoids the $O(d_k^3)$ matrix inversion at every step, but still incurs an $O(d_k^2)$ per-step overhead for maintaining the inverse.
Unlike the diagonal state used in our practical model, the Sherman-Morrison inverse recurrence is not a simple prefix-sum update, making chunk-parallel implementation less straightforward.
The full algorithm is shown in Algorithm~\ref{alg:ia-deltanet-sm}.


\begin{algorithm}[h]
\caption{IA-DeltaNet in recurrent form via Sherman--Morrison.}
\label{alg:ia-deltanet-sm}
\begin{algorithmic}[1]
\Require $\{\bm{k}_t,\bm{q}_t,\bm{v}_t,\beta_t\}_{t=1}^T$
\State $\bm{S}_0 \gets \bm{0}\in\mathbb{R}^{d_v\times d_k}$,
       \quad $\bm{P}_0 \gets \bm{I}\in\mathbb{R}^{d_k\times d_k}$ \Comment{$\bm{P}_t = \bm{\Sigma}_t^{-1}$}
\For{$t=1,\ldots,T$}
    \State $\bm{p}_t \gets \bm{P}_{t-1}\bm{k}_t$
    \State $s_t \gets \bm{k}_t^{\top}\bm{p}_t$
    \State $\bm{u}_t \gets \bm{p}_t / s_t$
    \State $\bm{e}_t \gets \bm{S}_{t-1}\bm{k}_t - \bm{v}_t$
    \State $\bm{S}_t \gets \bm{S}_{t-1} - \beta_t\,\bm{e}_t\,\bm{u}_t^{\!\top}$
    \State $\bm{P}_t \gets \bm{P}_{t-1} - \bm{p}_t\bm{p}_t^{\!\top} / (1 + s_t)$
    \State $\bm{o}_t \gets \bm{S}_t\bm{q}_t$
\EndFor
\State \Return $\{\bm{o}_t\}_{t=1}^T$
\end{algorithmic}
\end{algorithm}

\subsection{Chunk-parallel Diagonal IA-DeltaNet}
\label{appx:pseudo-chunk}

The implementation we use in practice mirrors the chunked DeltaNet forward pass of Flash Linear Attention,\footnote{\url{https://github.com/fla-org/flash-linear-attention/blob/main/fla/ops/delta_rule/naive.py}} replacing the write-side $\bm{k}_t$ with $\bm{u}_t$ and carrying the diagonal state $\bm{d}$ across chunks.
Since the diagonal state $\bm{d}$ is updated by an element-wise prefix sum of squared keys, it is compatible with chunk-parallel training via associative scans.


\small

\begin{verbatim}
def chunk_diagonal_ia_deltanet(q, k, v, beta, S, d, chunk_size):
    # q, k: (B, H, T, d_k), v: (B, H, T, d_v)
    # S: recurrent memory state, using FLA row-vector convention
    # d: diagonal past-key overlap state, shape (B, H, d_k)

    v_beta = v * beta[..., None]
    k_beta = k * beta[..., None]

    q, k, v_beta, k_beta = chunk(q, k, v_beta, k_beta, chunk_size)
    outputs = []

    for q_i, k_i, v_i, k_beta_i in zip_chunks(q, k, v_beta, k_beta):

        # Diagonal IA write directions within the chunk.
        k_sq = k_i * k_i
        k_sq_prefix = cumsum(k_sq, dim=time)
        d_before = d[:, :, None, :] + shift_right(k_sq_prefix)

        p = k_i / clamp(d_before)
        s = sum(k_i * p, dim=-1, keepdim=True)
        u_star = p / clamp(s)

        # Chunked DeltaNet transform with write-side K^T -> U_star^T.
        T = -tril(k_beta_i @ transpose(u_star), diagonal=-1)
        T = solve_triangular(I - T, I, lower=True, unitriangular=True)

        v_bar = T @ v_i
        k_bar = T @ k_beta_i          # read side remains key-based
        delta = v_bar - k_bar @ S

        # Intra-chunk output and recurrent state update.
        A = tril(q_i @ transpose(u_star))
        o_i = q_i @ S + A @ delta
        S = S + transpose(u_star) @ delta

        # Carry diagonal overlap state to the next chunk.
        d = d + k_sq_prefix[:, :, -1]

        outputs.append(o_i)

    return concat(outputs, dim=time), S, d
\end{verbatim}


\end{document}